\documentclass[]{llncs}
\usepackage{arabtex}
\usepackage{utf8}
\usepackage[utf8x]{inputenc}
\usepackage[T1]{fontenc}
\usepackage{graphicx}
\begin{document}
%

\title{SentiALG: Automated Corpus Annotation for Algerian Sentiment Analysis}

%

%
\author{Imane Guellil\inst{1,2} \and Ahsan Adeel\inst{3} \and Faical Azouaou\inst{2} \and Amir Hussain \inst{3} }
 \institute{
            Ecole Superieure des sciences appliquées d'Alger ESSA-alger \email{i.guellil@essa-alger.dz}
            \and 
            Laboratoire des Méthodes de Conception des Systèmes (LMCS), 
            Ecole nationale Supérieure d’Informatique,BP 68M, 16309, Oued-Smar, Alger, Algérie \email{i\_guellil@esi.dz, f\_azouaou@esi.dz}
            \and
            Institute of Computing science and Mathematics, School of Natural Sciences University of Stirling Stirling UK \email{ahsan.adeel@stir.ac.uk, ahu@cs.stir.ac.uk}}

\maketitle              
\begin{abstract}
Data annotation is an important but time-consuming and costly procedure. To sort a text into two classes, the very first thing we need is a good annotation guideline, establishing what is required to qualify for each class.  In the literature, the difficulties associated with an appropriate data annotation has been underestimated. In this paper, we present a novel approach to automatically construct an annotated sentiment corpus for Algerian dialect (A Maghrebi Arabic dialect). The construction of this corpus is based on an Algerian sentiment lexicon that is also constructed automatically. The presented work deals with the two widely used scripts on Arabic social media: Arabic and Arabizi. The proposed approach automatically constructs a sentiment corpus containing 8000 messages (where 4000 are dedicated to Arabic and 4000 to Arabizi). The achieved F1-score is up to 72\% and 78\% for an Arabic and Arabizi test sets, respectively. Ongoing work is aimed at integrating transliteration process for Arabizi messages to further improve the obtained results.


\end{abstract}
%
%
%
\section{Introduction}
Sentiment analysis is defined as an interdisciplinary domain among the natural language processing (NLP), artificial intelligence (AI), and text mining \cite{guellil2015social}. To determine whether a document or a sentence
expresses a positive or negative sentiment, three main approaches
are commonly used: the lexicon based approach \cite{taboada2011lexicon}, machine learning (ML) based approach \cite{maas2011learning} and a hybrid approach \cite{khan2015combining}.
English  has  the greatest 
number  of  sentiment  analysis  studies
, while  research  is  more 
limited  for  other  languages  including  Arabic and its dialects \cite{alayba2017arabic}. 

ML based sentiment analysis requires an annotated data. The lexicon based approach needs an annotated sentiment lexicon (containing the valence and/or intensity of its terms and/or expressions). One of the majors problems related to the treatment of Arabic and its dialect is the lack of resources. Other dominant problems include the standard  romanization (called Arabizi) that Arabic speakers often use in social  media. Arabizi uses Latin alphabet, numbers, punctuation for writing an Arabic word (For example the word "mli7", combining between Latin letters and numbers, is the romanized form of the Arabic word "\<mly.h>" meaning "good"). To the best of our knowledge, limited work has been conducted on sentiment analysis of Arabizi and it is dedicated to Arabic and not to its dialect \cite{duwairi2016sentiment}. However, not much work has been conducted on sentiment analysis of Algerian Arabizi. 

To bridge the gap, this paper proposes an approach that automatically construct a sentiment lexicon for a Magheribi dialect (i.e. Algerian dialect). Based on the constructed lexicon, we automatically annotate a sentiment corpus into positive and negative. To validate the build corpus, we applied a set of classifiers and tested our corpus on two different test sets: internal (which is a part of the constructed corpus) and external (which represent a set of messages that we manually annotated). However, the general experimental results have shown  better performance with Arabic test sets which is attributed to the complexity of Arabizi.

This paper is organized as following: 
Section 2 presents the related work on sentiment analysis by focusing on the work done on Arabic and its dialects. Section 3 presents our approach and the different parts that composed it. Section 4 presents the different results that we collected in this study. Section 5 presents conclusion containing some opening for our futures works.

\section{Arabic Sentiment Analysis}
\subsection{Lexicon-based approaches}
A lexicon of 120,000 Arabic terms is build  in \cite{al2015lexicon}, following infinitives collection, transliteration to English, and exploitation of English lexicon to determine the valence and intensity of each word. Another large lexicon has been constructed in \cite{badaro2014large}. It contains 157 969 synonymous and 28 760 lemmas. To build this big dataset, the authors combined several Arabic resources including English WordNet, Arabic WordNet, English SentiWordNet, Standard Arabic Morphological Analyzer (SAMA) . In \cite{mohammad2013crowdsourcing}, the authors develop a lexicon of sentiment containing 14 182 English uni\-gram classified into positive or negative using the "Mechanical Turk of Amazon"\footnote{https://www.mturk.com/}. This lexicon is then translated into 40 languages including MSA. The authors in \cite{abdulla2014automatic} studied three lexical construction techniques including one manual and two automatics. In addition, a SA tool was developed within this work. Experiments showed that the use of a lexicon containing 16 800 words (created by integrating three techniques, so one manual and the two others automatic) gives the best results. 
In \cite{mataoui2016proposed}, the authors manually construct a lexicon of sentiment starting with an existing Arabic and Egyptian lexicon. They analyzed messages containing MSA as well as DALG. To answer to the morphological characteristics of this language and dialect, the authors used the lemmatization tool "khoja" (developed for MSA).

Most of the proposed lexicon construction methods are based on three:
1) manual; 2) automatic translation and 3) annotated corpus.In this paper, we exploited the second technique to construct our Algerian sentiment lexicon. 
\subsection{Machine learning based approaches}
Supervised approaches essentially depends on the existence of annotated data. Among the corpora presented in the literature, we cite: OCA \cite{rushdi2011oca}, AWATIF \cite{abdul2012awatif},LABR \cite{aly2013labr}, TSAC \cite{medhaffar2017sentiment}, AraSenTi-Tweet \cite{al2017arasenti}. OCA contains 500 Arabic comments (250 positive and 250 negative), manually preprocessed, then segmented, and root extracted with a tool dedicated to Arabic. AWATIF is a multi-genre corpus containing 10 723 sentences in Arabic manually annotated in objective and subjective sentences. Then annotation of the subjective sentences in positive, negative or neutral. LABR contains 63 257 Arabic comments annotated with stars ranging from 1 to 5 by users. The authors considered positive comments containing 4 or 5 stars, negative ones containing 1 or 2 stars and neutral ones containing 3 stars. TSAC contains 17 060 comments (including 8215 positive and 8845 negative) in Tunisian dialect annotated manually. AraSenTi-Tweet contains 17 573 Saudi tweets, manually annotated into four classes (positive, negative, neutral and mixed). 

Almost all works are based on the constructed corpus to classify sentiment (by using classification algorithm). The most used classification Algorithm are: Support Vector Machine (SVM) and Naive Bays(NB). However, most of the aforementioned works suffer from: manual annotation, almost all resources are not publicly available and constructed corpora are not dedicated to DALG. To the best of our knowledge, one work only has been done on Arabizi sentiment analysis \cite{duwairi2016sentiment}. However, it is not focus on Algerian Arabizi. Emergent works have been done on Algerian Arabizi treatment \cite{guellilbilingual,guellil2017asda,guellil2017arabizi,guellil2017comparison,guellil2016arabic} but no one concentrate on sentiment analysis

\section{Contribution}
In this paper, we present an approach for sentiment analysis of DALG messages. This approach is based on an annotated corpus that we constructed automatically. The construction of this corpus is based on a sentiment lexicon (that we also constructed automatically) and on a sentiment algorithm (handling DALG characteristics).  Figure 1 presents the proposed sentiment analysis approach.  The proposed approach constitutes three general steps: 1) Automatic construction of DALG sentiment lexicon. 2) Polarity calculation of DALG messages and 3) Sentiment classification of DALG messages. The three steps are comprehensively explained in subsequent sections.
\begin{figure} 
\centering\includegraphics[scale=0.5]{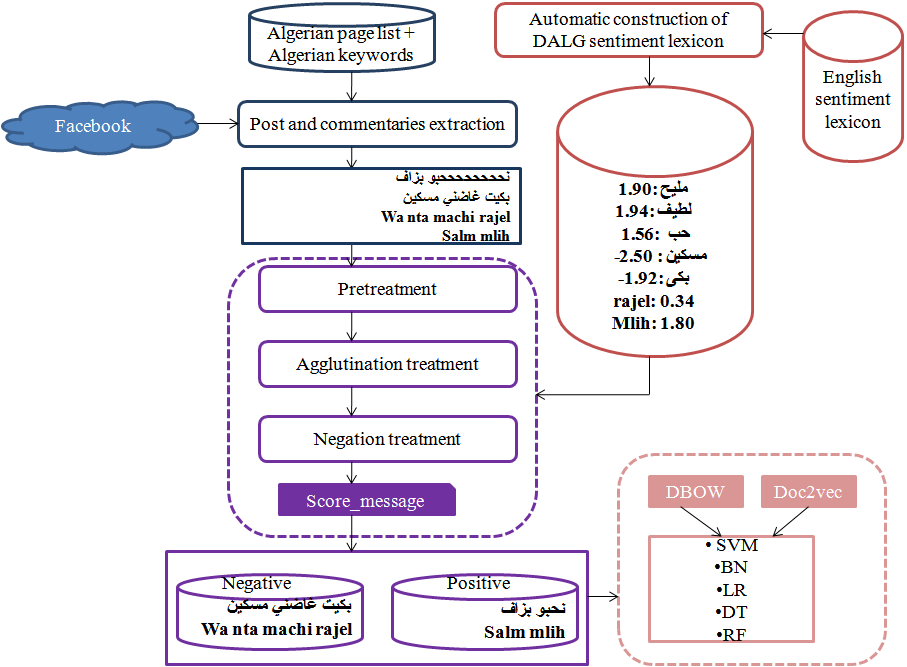}
  \caption{A general architecture of our approach for sentiment analysis DALG}
\end{figure}
\subsection{Automatic construction of DALG sentiment lexicon}
Our approach receives a lexicon of sentiments in English as input. Each word in this lexicon is translated using a translation API (\footnote{https://glosbe.com/en/arq/excellent}). The specificity of this API is that the translation is performed  by ordinary users native of the DALG. The same score is assigned to all collected words. For example, the English word 'excellent' (with a score of +5) gives different DALG words such as: '\<bAhy>' baAhiy, '\<l.tyf>' lTiyf, '\<mly.h>', mliyH, etc. All these words receive a score of +5, similarly to the word 'excellent'.
However, a word in DALG is associated with several words in English and can therefore have several scores. For example the word \<mly.h> mliyH, 'good' is associated with several English words such as: excellent, generous, delicious, etc. The word 'excellent' has a score equal to (+5); the word 'generous' has the score (+2) and the others words have different scores. Therefore we extract, within this part, all the words in DALG and calculating their scores. Concerning the calculation of the score, we take the average of the scores of all the English words to which our word in DALG is associated. 
\subsection{Polarity calculation of DALG messages}
The goal of this step is to automatically annotate a set of Facebook messages (extracted from Algerian pages) as positive and negative. For example, the sentence: '\<n.hbw bzAf>' translated into 'I love him a lot' should recognized as positive  and the sentence 'wa nta machi rajel' translated into 'and you are not man' should recognized as negative. To correctly annotate this sentences and others, we need to proceed to a set of treatments: 1) Pretreatment of the messages. 2) Agglutination treatment. 3) Negation treatment.
\subsubsection{Pretreatment of the messages}
\begin{itemize}
\item Deleting of the repeated messages in the corpus to keep only one occurrence of each message.
\item Deleting of exaggerations, for example the word \<n.h.h.h.h.h.h.hb> is transformed into: \<n.hb> and nhhhhhab is transformed into nhab. The different repetitions of the different letter '\<.h>' and 'h' are removed to keep a single occurrence.
\item Deleting of the '\#' character and spaces of the different punctuations '.,!,?' of the related word .
\item Deleting of consecutive whites spaces as well as Tatweel ('--') within Arabic characters.
\end{itemize}
\subsubsection{Agglutination treatment}
We first form all the possible n-gram in the messages. For example, the second message ('\<bkyt .gA.zny mskyn>') gives us 3 uni-gram ('\<bkyt,  .gA.zny> and  \<mskyn>'), 2 bi-gram (\<bkyt .gA.zny, .gA.zny mskyn>') and one tri-gram      
('\<bkyt .gA.zny mskyn>'). Then, we look for each n-gram in the lexicon and add the score of the finding ones into the global score (if they are not proceeded by a negation).
If no n-gram is found, we extract stem from each word and look for it in the lexicon. To extract the DALG stems, we define a set of prefixes and suffixes (Personnel pronouns, complement pronouns, feminine pronouns, plural pronouns, etc) of DALG. For example: The stem of the word '\<n.hbw>' is '\<.hb>' because the two letters '\<n>' and '\<w>' are respectively represent a prefix and a suffix in DALG and the stem '\<.hb>' is recognized in our lexicon. 
However, some words which are conjugated into the past like '\<bkyt>' to be transformed before being recognized. So, the stem of '\<bkyt>' is '\<bkY>'. To recognize it, we have to extract the part without affixes (so '\<bk>' and add the letter '\<Y>' at the end).
\subsubsection{Negation treatment}
Negation analysis is an important research challenge for all languages. Nevertheless this challenge is accentuated in the case of Arabic and its dialects where the negation is usually attached to the word as well as the different pronouns. Users can use negation in different ways, for example the word \<mAn.hbkm^s> can be written in the following way: \<mA n.hbkm^s> or \<mAn.hbkm ^s> or \<mA n.hbkm ^s>. Therefore, we notice that negation can be attached to or separated from terms. 
We have found, however, that in most cases negation does not only affect the preceding word  but also some other words in the sentence. Once a prefix or negation suffix is detected, we reverse the score of the words succeeding this negation (multiplying the score by (-1)).

After calculating the score of each messages, we annotate it as positive (if its score is bigger than 0), and negative in the other case (so if its score is smaller that 0). Finally, we are able to automatically annotate the extracted corpus.
\subsection{Sentiment classification of DALG messages}
In this paper, we propose to compare different shallow classification models. We used two vectorization techniques: Bag of Words (BOW) and document embedding where we rely on the Doc2vec algorithm  presented within \cite{le2014distributed}. For Doc2vec, we apply the two methods presented in \cite{le2014distributed}: 1) Distributed Memory Version Of Paragraph Vector (PV-DM) et 2) Distributed Bag of Words Version of Paragraph Vector (PV-DBOW). We also use the implementation merging these two methods. In the method (PV-DM), the paragraph vector (document or sentence) is concatenated to the word vectors in order to predict the next word within a text window. Unlike this method, (PV-BOW) ignores the context of the words within the inputs and this to force the prediction of these words randomly focusing on the paragraph vector.
For the classification part, we use different classifiers: 1) Support Vector Machine (SVM). 2) Naive Bayes (NB). 3) Logistic regression (LR). 4) Decision Tree (DT) and 5) Random Forest (RF).


\section{Experimentations and results}
\subsection{Constructed resources}
In this work we constructed three kind of resources:  1) An Algerian sentiment lexicon (containing words in both Arabic and Arabizi). 2) A monolingual Algerian dialect corpus. 3) An annotated Algerian sentiment corpus.

For the construction of lexicons, we used SOCAL (an English sentiment lexicon) presented in \cite{taboada2011lexicon}. SOCAL contains 6769 terms 
whose sentiment is labeled between (-1, -5) for negative terms and between (+1, +5) for positive terms. However only 3968 terms in English have been recognized and translated by the Glosbe API. After the extraction of Algerian dialect terms we obtain a lexicon containing 4873 words annotated from (-5 to +5), where 2390 are in Arabic and 2483 are in Arabizi.

We use Socialbekers website\footnote{https://www.socialbakers.com/statistics/facebook/pages/total/algeria/} to collect the name of the 226 most famous Algerian Facebook pages including Ooredoo, HamoudBoualem, Algeria Telecom, Ruiba, etc. We also extracted some strong dialectal Algerian keywords from PADIC corpus\cite{meftouh2015machine} (for example, \<yfr.h> which means "he is happy", \<AlkdAb > which means "a lier") .Then, we use RestFB\footnote{http://restfb.com/} API implemented with JAVA to extract all post and commentaries present in the target pages and present in others pages but containing Algerian keywords. At the end, we were able to collect \textbf{15,407,910}           messages where \textbf{7,926,504} are in Arabic and \textbf{3,976,700} are in Arabizi.

Based on the constructed lexicon, on the extracted corpus and on the sentiment analysis algorithm that we developed, we are able to automatically annotate a sentiment corpus. Then we randomly form a training corpus containing 8000 messages (where 4000 messages are in Arabic  and 4000 messages are in Arabizi with 2000 positives messages and 2000 negative. After, this corpus is divided into three part (train, dev and test) for internal experiment and (train, dev) for external experiment. The dev set is used only for deep classification. For shallow classification, we use the entire annotated corpus as training. 

\subsection{Experimental results}
The constructed automatically annotated sentiment corpus has been critically analyzed  by applying shallow algorithms. Difficulties to deal with a non-resourced dialect (such as Algerian dialect which uses two different alphabets) have also been highlighted.Table 1 presents the performance of difference classification algorithms in terms of Accuracy (Acc)) and F1-score (F1) for each vectorization method (BOW, Doc2vec).
\begin{table} 
\centering\includegraphics[scale=0.44]{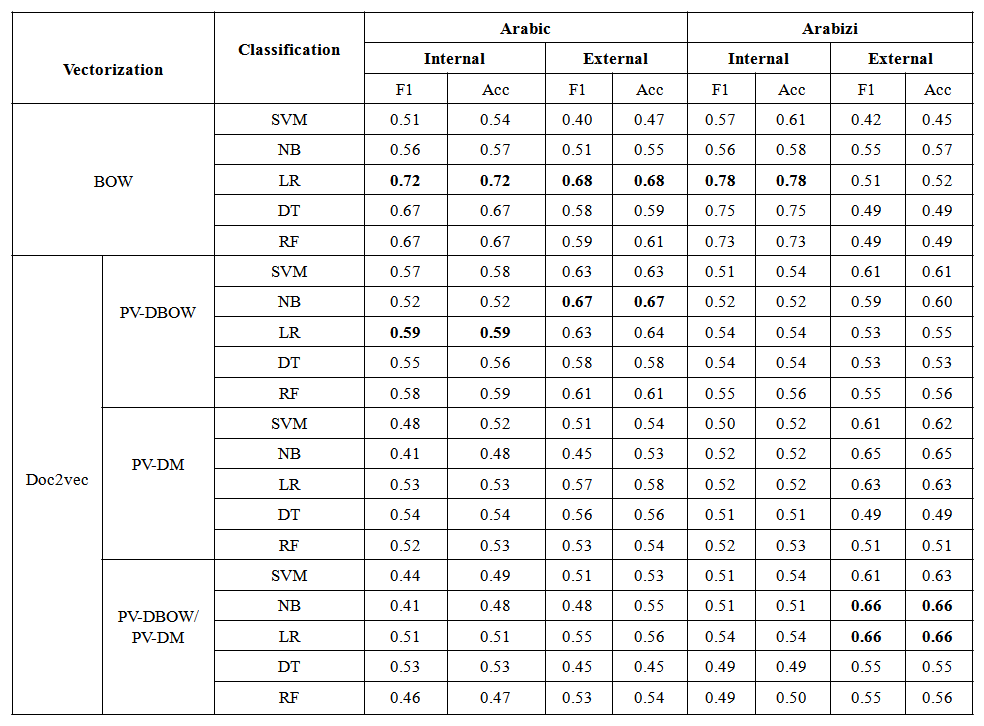}
  \caption{DALG Sentiment analysis results}
\end{table}

Based on simulations and analysis,  our three major observations are: 1) The results with  Arabic sets are better than Arabizi sets in most of the cases. 2) The results on internal data sets are better than external ones. 3) Bow vectorization gives better results than Doc2vec.
These results are principally related to five facts and could substantially be improved by handling these lack: 
\begin{itemize}
\item Arabizi is very complex and one term could have different writing manner (sometimes more than 100 manners). handling it with one lexicon is almost impossible. To address this issues, we could enrich the used lexicon by all variation of Arabizi word. We  also propose to integrate a transliteration module to transform Arabizi into Arabic before analyzing the sentiment of Arabizi messages.
\item Our lexicon contains lots of noise. For example the different words: 1) "\<b, w, m`>", etc (with) have a negative polarity when these words are neural.2) The word ("\<rb>" meaning "god" has a positive polarity in Arabic when the word "lah" representing a part of "alah" in Arabizi (meaning god too) has a negative polarity. To address this problem, a manual review of  lexicons could be used to increase the precision of annotation.
\item Other parameters than score involved in the annotation process such as: number of positives and negatives words, length of the sentence, the comparison of score to other threshold than 0, etc. The integration of these parameters to our algorithm and testing and the influence of each one on the annotation process could considerably improve t,he results.
\item Some irregular plurals such as \<mlA.h> in arabic and "mlah" in Arabizi are not recognize by our algorithm which is only based on soft stemming. The proposition of a stemmer tool dedicated to DALG could improve the annotation process.
\item The vectorization used techniques are complementary. Hence, some messages are recognized by using DBOW and are not recognized by using Doc2vec and vice versa. The combination between the different vectorization technique will considerably improve the results. 
\section{Conclusion and Perspectives}
In this paper, we present a novel approach to automatically construct an annotated sentiment corpus for Algerian dialect (A Maghrebi Arabic dialect). The construction of this corpus is based on an Algerian sentiment lexicon that is also constructed automatically. The presented work deals with the two widely used scripts on Arabic social media: Arabic and Arabizi. The proposed approach automatically constructs a sentiment corpus containing 8000 messages (where 4000 are dedicated to Arabic and 4000 to Arabizi). The achieved F1-score is up to 72\% and 78\% for an Arabic and Arabizi internal test sets, and up to 68\% and 66\% for an Arabic and Arabizi internal test sets respectively. 

This study represents the baseline for our future work where we plan to augmenting the
lexicon size based on embedding algorithm such as word2vec.  Additionally, enlarging the dataset with focusing on more annotated words will provide much better results. One thing more, one arabizi word could have many different writing manners. This phenomena leads to misinterpretation and consequently a wrong polarity classification. Hence, proceeding to transliterate arabizi messages absolutely improve the results.

We also plan to extend our work for handling handwriting Arabic based on the work proposed in \cite{alkhateeb2011offline,alkhateeb2011performance,alkhateeb2009component}.
\section*{Acknowledgment}
Amir Hussain and Ahsan Adeel were supported by the UK Engineering and Physical Sciences Research Council (EPSRC) grant No.EP/M026981/1.

\end{itemize}

\bibliography{biblio}
\bibliographystyle{splncs04}

\end{document}